\documentclass[runningheads]{llncs}
\usepackage{hyperref}
\hypersetup{
    colorlinks=true,
    linkcolor=blue,
    filecolor=blue,      
    urlcolor=blue,
    citecolor=blue
}
\usepackage{cite}
\usepackage{graphicx}
\usepackage{caption}
\usepackage{subfigure}
\usepackage{indentfirst} 
\usepackage{amsmath}
\usepackage{bm}
\usepackage{multirow}
\usepackage{booktabs}
\usepackage{graphicx}
\usepackage[table,xcdraw]{xcolor}
\usepackage[section]{placeins}

\begin{document}
\title{Learning Robust Representation for Joint Grading of Ophthalmic Diseases via Adaptive Curriculum and Feature Disentanglement}
\titlerunning{Learning Robust Representation for Joint Grading of Ophthalmic Diseases}
% % If the paper title is too long for the running head, you can set
% % an abbreviated paper title here

\author{Haoxuan Che\inst{1,3} \and Haibo Jin\inst{1} \and Hao Chen\inst{1,2,3}} 
%index{Che, Haoxuan}
%index{Jin, Haibo}
%index{Chen, Hao}

\authorrunning{H. Che, et al.}
% % % First names are abbreviated in the running head.
% % % If there are more than two authors, 'et al.' is used.

\institute{Department of Computer Science and Engineering \and Department of Chemical and Biological Engineering \and Center for Aging Science \\
The Hong Kong University of Science and Technology, Kowloon, Hong Kong \\
\email{\{hche, hjinag, jhc\}@cse.ust.hk}}

\maketitle              % typeset the header of the contribution
\begin{abstract}
Diabetic retinopathy (DR) and diabetic macular edema (DME) are leading causes of permanent blindness worldwide.
Designing an automatic grading system with good generalization ability for DR and DME is vital in clinical practice.
However, prior works either grade DR or DME independently, without considering internal correlations between them, 
or grade them jointly by shared feature representation, yet ignoring potential generalization issues caused by difficult samples and data bias. 
% These neglects hinder models learning robust features and thus harm generalization abilities.
Aiming to address these problems, we propose a framework for joint grading with the \textbf{d}ynamic difficulty-\textbf{a}ware \textbf{w}eighted loss (DAW) and the \textbf{d}ual-str\textbf{e}am disen\textbf{t}angled le\textbf{a}rning ar\textbf{ch}itecture (DETACH). 
Inspired by curriculum learning, DAW learns from simple samples to difficult samples dynamically via measuring difficulty adaptively.
DETACH separates features of grading tasks to avoid potential emphasis on the bias.
With the addition of DAW and DETACH, the model learns robust disentangled feature representations to explore internal correlations between DR and DME and achieve better grading performance.
Experiments on three benchmarks show the effectiveness and robustness of our framework under both the intra-dataset and cross-dataset tests.
% \keywords{Diabetic retinopathy \and Diabetic macular edema \and Joint grading}
\end{abstract}
\section{Introduction}
\begin{figure}[t]
    \centering
    \includegraphics[width=0.95\textwidth]{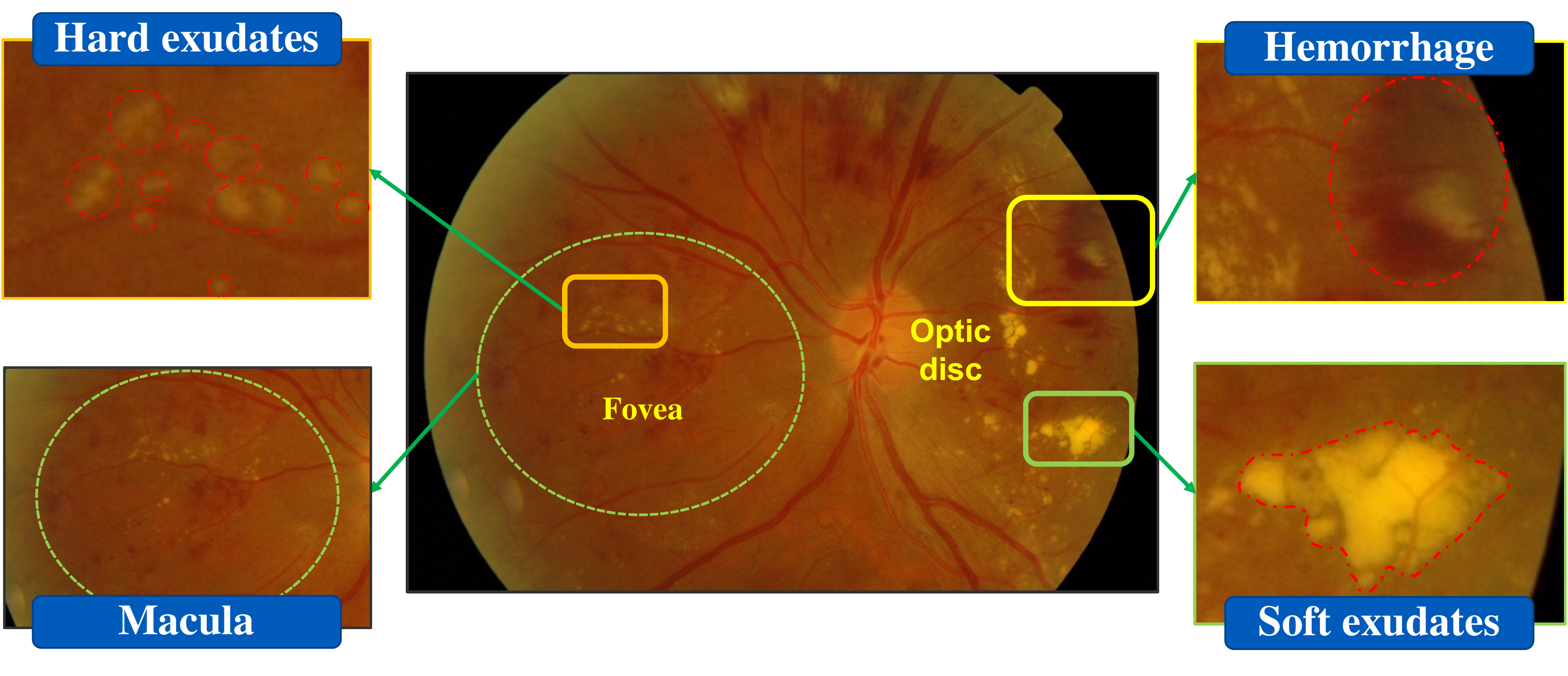}
    \caption{Symptoms of DR and DME are different in fundus images\cite{cho2018idf,decenciere2014feedback,das2015diabetic, porwal2018indian}. Physicians grade DR via soft and hard exudates, hemorrhage, etc., yet determine DME via occurrences of hard exudates and the shortest distances of hard exudates to the macula center.} \label{intro}
\end{figure}

Diabetic retinopathy (DR) and diabetic macular edema (DME) are complications caused by diabetes, which are the most common leading cause of the visual loss and blindness worldwide \cite{cho2018idf}.
It is vital to classify stages of DR and DME in clinical practice because treatments are more effective for those diseases at early stages, and patients could receive tailored treatments based on severity.
Physicians grade DR into multiple stages according to the severity of retinopathy lesions like hemorrhages, hard and soft exudates, etc., while they classify DME into three stages via occurrences of hard exudates and the shortest distances of hard exudates to the macula center \cite{decenciere2014feedback,porwal2018indian}, as shown in Fig.\ref{intro}. 

Recently, large progress has been made by deep learning based methods on the grading of DR \cite{zhou2018multi,he2020cabnet,liu2020green,tian2021learning} and DME \cite{ren2018diabetic,syed2018fundus}.
For example, He et al. \cite{he2020cabnet} proposed a novel category attention block to explore discriminative region-wise features for each DR grade,  
and Ren et al. \cite{ren2018diabetic} presented a semi-supervised method with vector quantization for DME grading. 
However, DR and its associated DME were treated as separate diseases in these methods.
Later, several works proposed to conduct the grading jointly \cite{gulshan2016development, krause2018grader, li2019canet}.
Among them, Gulshan et al. \cite{gulshan2016development}, and Krasuse et al. \cite{krause2018grader} only leveraged the relationship implicitly by treating them as a multi-task problem.
Subsequently, Li et al. \cite{li2019canet} explored the internal correlation between DR and DME explicitly by designing a cross-disease attention network (CANet).

% Although existing joint grading efforts show fruitful effectiveness, they still ignore two vital DR and DME joint grading issues.
Despite the fruitful effectiveness of existing works on DR and DME joint grading, two vital issues remain unsolved.
First, there is no proper solution for difficult samples, even though they have raised challenges for DR and DME grading tasks.
For example, some previous works re-modify the 4-class DR grading task as a binary classification due to challenging samples \cite{li2019canet,wang2017zoom,vo2016new}.
In clinical settings, however, fine-grained grading is necessary due to the need for tailored treatments to avoid visual loss of patients.
% Second, they adopt entangling feature representations for DR and DME grading, which may emphasize the potential bias of models caused by class imbalance and stereotyped disease context
Second, the entangling feature representations are prone to emphasize the potential bias caused by the class imbalance and stereotyped disease context \cite{chu2021learning}, where severe DR always accompanies DME in existing public datasets \cite{decenciere2014feedback, porwal2018indian}, however, DME can occur at any stage of DR \cite{das2015diabetic}. 
% Such emphasized potential bias can lead to subpar performance on cross-domain data.
The bias can harm the generalization ability of the model, leading to subpar performance on cross-domain data.
% This is particularly true for under-represented classes, where a lack of diversity in the data exacerbates the tendency.
Models with robust generalization ability are especially significant for clinical application \cite{geirhos2020shortcut,chen2019robust}. 
However, existing works do not explore how to avoid the potential bias and mitigate the degeneration of generalization performance.

Therefore, we propose a difficulty-aware and feature disentanglement-based framework for addressing difficult samples and potential bias to jointly grade DR and DME.
Firstly, inspired by curriculum learning (CL) \cite{bengio2009curriculum}, we propose the \textbf{d}ynamic difficulty-\textbf{a}ware \textbf{w}eighted loss (DAW) to learn robust feature representations.
% DAW could evaluate the difficulties of samples adaptively, and weight them via consistency between model predictions and ground-truth labels.
DAW weights samples via the evaluation of their difficulties adaptively by the consistency between model predictions and ground-truth labels.
Moreover, it focuses on learning simple samples at early training, and then gradually emphasizes difficult samples, like learning curriculum from easy to challenging.
Such a dynamic and adaptive curriculum helps models build robust feature representations and speed up training \cite{bengio2009curriculum}.
Meanwhile, given the success of feature disentanglement in generalization \cite{trauble2021disentangled,montero2020role}, we design a \textbf{d}ual-str\textbf{e}am disen\textbf{t}angled le\textbf{a}rning ar\textbf{ch}itecture (DETACH).
DETACH prevents potential emphasized bias by disentangling feature representations to improve the robustness and generalization ability of models.
It enables encoders to receive supervision signals only from their tasks to avoid entangling feature representations while retaining the ability to explicitly learn internal correlations of diseases for the joint grading task. 
Experiments show that our framework improves performance under both intra-dataset and cross-dataset tests against state-of-the-art (SOTA) approaches.

\begin{figure}[t]
    \centering
    \includegraphics[width=\textwidth]{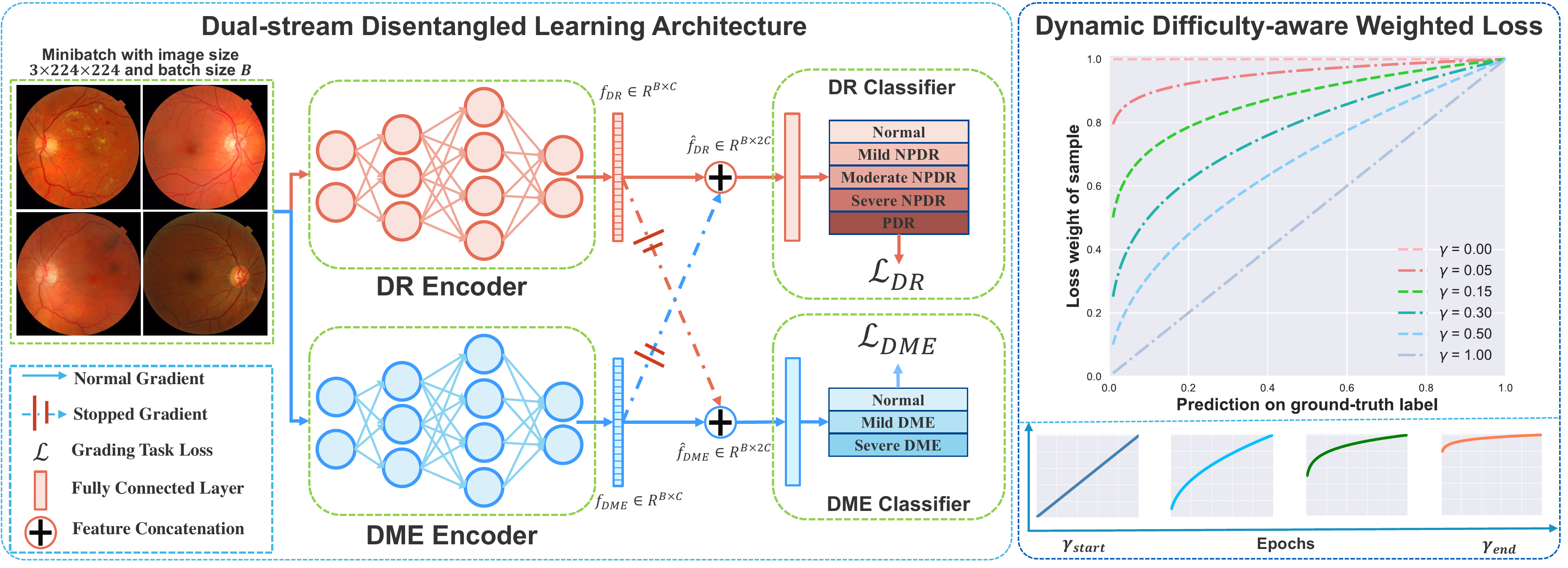}
    \caption{The overview of our proposed framework. The framework uses two encoders to extract disentangled features $f_{DR}$ and $f_{DME}$ for DR and DME grading tasks, respectively. 
%     Features flow into classifiers to learn correlations between tasks, yet gradients are just back-propagated to their upstream encoders. 
The concatenated features, $\hat{f}_{DR}$ and $\hat{f}_{DME}$, flow into classifiers to learn correlations between tasks, while the model stops gradients of DR features back-propagating to the DME encoder, and vice versa.
The dynamic difficulty-aware weighted loss (DAW) weights samples adaptively via predictions on true labels with $\gamma$ adjusting weights dynamically during training.} \label{pipeline}
\end{figure}

\section{Methodology}
An overview of our framework is shown in Fig.\ref{pipeline}.
Our framework disentangles feature representations of DR and DME via cutting off gradient back-propagation streams,
and learns from simple to difficult samples via controlling the difficulty-aware parameter.
% In this section, we will first demonstrate a preliminary analysis of the Messidor dataset \cite{decenciere2014feedback} in order to show clues of difficult samples and generalization performance degeneration, and then introduce the proposed methods to address these issues.
In this section, we first show clues of difficult samples and generalization performance degeneration through a preliminary analysis on the Messidor dataset \cite{decenciere2014feedback}, and then introduce the proposed methods to address these issues.
% This section will first illustrate a preliminary analysis on the Messidor dataset \cite{decenciere2014feedback} to show clues of difficult samples and generalization performance degeneration, and then introduce proposed components DAW and DETACH for addressing these issues.
% as : how to weight samples adaptively, how to learn as curriculum learning, and how to disentangle feature representation.

\begin{figure}[t]
    \centering
    \includegraphics[width=1\textwidth]{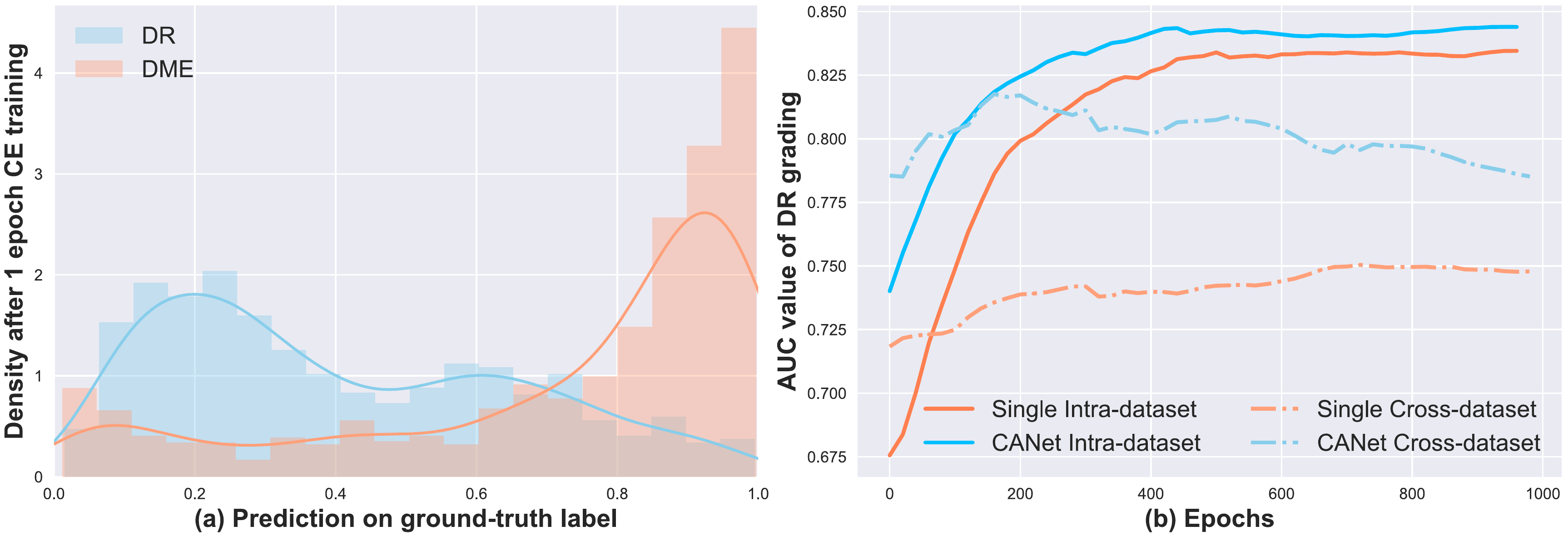}
    \caption{Fig.(a) is a probability distribution histogram of model predictions on ground-truth labels after one epoch cross-entropy loss (CE) training, which shows divergent learning difficulties among samples and between tasks. 
Fig.(b) shows increasing performance in the intra-dataset test yet different trends in the cross-dataset test, implying potential emphasized bias and generalization performance degeneration.} \label{analysis}
\end{figure}

\subsubsection{Preliminary analysis.}
Fig.\ref{analysis}(a) implies divergent difficulties among samples and between grading tasks, as also reported in \cite{sanchez2011evaluation}, which claims a phenomenon of low agreement on grading for experts even adhering to a strict protocol.
Such grading difficulty may be introduced by the ambiguity of samples near decision boundaries \cite{toneva2018empirical}.
Unlike standard samples, difficult samples with ambiguity require special learning strategies \cite{yang2020metric}.
Besides, as shown in Fig.\ref{analysis}(b), an entangling representation-based model, CANet, improves its DR grading performance in the intra-dataset test yet performs continual-decreasingly in the cross-dataset test.
On the contrary, a vanilla ResNet50 trained only for DR grading (denoted as Single) with cross-entropy loss (CE) improves performance in both tests.
A possible explanation \cite{chu2021learning,trauble2021disentangled} is that the model with entangled features may emphasize potential class imbalance and stereotyped correlations with DR and DME during intra-domain training, and thus suffers from performance decrease on cross-domain data.
Our preliminary analysis implies the existence of difficult samples and generalization ability degeneration in DR and DME joint grading. 
Thus, reasonable learning strategies and disentangled feature representations are necessary for DR and DME grading.

\subsubsection{Weighting samples via difficulty adaptively.}
As illustrated previously, CL can help handle difficult samples \cite{bengio2009curriculum}. 
However, one critical puzzle for CL is how to measure the difficulties of samples reasonably.
% Instead of needing extra models, variables, or the knowledge of experts, we alternatively use predictions on labels to adaptively evaluate the difficulties of samples for the model.
Instead of requiring extra models or the knowledge of experts to measure difficulty, we alternatively use consistency between model predictions and labels to evaluate the difficulties of samples for the model adaptively.
It is inspired by the implicit weighting scheme of CE \cite{zhang2018generalized}.
For notation convenience, we denote $y$ as the one-hot encoding label of the sample, $p$ as the softmax output of the model, and $p_t$ as the probability of the true class where the sample belongs. 
We define the loss as $\text{CE}(p,y) = -log(p_t)$ and show its gradient as:
\begin{equation}
% 		\frac{\partial\mathcal{L}(f(x; \bm{\theta}), y)}{\partial\bm{\theta}} = 
        \frac{\partial \text{CE}(p,y)}{\partial \theta} = 
        % \begin{cases}
        -\frac{1}{{p_t}}\nabla_{\bm{\theta}} p_t %& \text{for CE} \\
        % -\nabla_{\bm{\theta}} p_t & \text{for MAE}
        % \end{cases} 
        \label{grad}
\end{equation}
where $\theta$ denotes parameters of the model.
As  Eq.\ref{grad} shows, CE implicitly weights more on the samples whose model predictions are less congruent with labels, i.e., those difficult samples with smaller $p_t$ and hence larger $1/p_t$, than samples whose model predictions are more consistent with labels, i.e., simple samples \cite{zhang2018generalized}.

Inspired by the implicit effects of $p_t$ on gradients of samples shown in Eq.\ref{grad}, we design a d\textbf{y}namic difficulty-\textbf{a}daptive \textbf{w}eighted loss (DAW) to explicitly mitigate the enhancement on difficult samples.
Specifically, we propose to add a difficulty-adaptive weight $\alpha^\gamma$ with a tunable difficulty-aware parameter $\gamma \in [0,1]$ to CE, and we define DAW as:
\begin{equation}
	    	\mathcal{L}_{DAW}(p, y, \gamma) = -\alpha^\gamma log(p_t) \\
\end{equation} 
where $\alpha$ is the numeric value of $p_t$, and thus $\alpha^\gamma$ also acts on gradients directly.
DAW weights samples by difficulty adaptively measured via $p_t$, which is continually updated during the training process.
Such adaptive consideration of the keep-evolving model delivers a proper curriculum, where difficulties of samples are continual-changing for the model \cite{jiang2018mentornet}.
Moreover, DAW can emphasize or reduce the effects of $p_t$ via controlling $\gamma$.
As shown in the right of Fig.\ref{pipeline}, 
larger $\gamma$, i.e., more sensitive difficulty-awareness, means that the model applies smaller weights on difficult samples, and vice versa.

\subsubsection{Learning from simple to difficult dynamically.}
DAW learns from simple to difficult samples by setting $\gamma$ initially as $\gamma_{start}$ and gradually decreasing it to $\gamma_{end}$, and the range $[\gamma_{end},\gamma_{start}]$ denotes the dynamic difficulty-aware range.
A larger $\gamma_{start}$ means that the model learns less on difficult samples at the beginning, and a smaller $\gamma_{end}$ means that the model treats samples more equally in the end. 
Such learning dynamics help the model to establish common features first, then try to learn distinctive features via its current knowledge during the training, and finally build robust feature representations \cite{bengio2009curriculum}.
After reaching $\gamma_{end}$, $\gamma$  stops decreasing and the model continues to learn with $\gamma = \gamma_{end}$. 
In this paper, we adopt a  linear decreasing strategy for $\gamma$, i.e., $\gamma$ decreases equally every epoch, to verify the effectiveness of DAW in the experiment section, but more sophisticated algorithms can be used to boost the performance further.

Compared to existing CL methods, DAW handles samples adaptively and dynamically.
To be specific, the existing works need extra networks or expert knowledge to evaluate the difficulty of samples as \cite{kumar2010self,jiang2018mentornet} or sample difficulties are usually predefined and fixed during training as \cite{bengio2009curriculum}. 
In contrast, for one thing, DAW uses continual-updating predictions to measure sample difficulties for the model at that moment adaptively; for another, decreasing $\gamma$ focuses more on learning difficult samples than before dynamically.
As a result, DAW could set a reasonable and effective curriculum for the model.
% via such adaption and dynamics.

\subsubsection{Disentangling feature representations of DR and DME.}
To improve the generalization ability of our model, we propose a feature disentanglement module, named  \textbf{d}ual-str\textbf{e}am disen\textbf{t}angled le\textbf{a}rning ar\textbf{ch}itecture (DETACH), shown in Fig.\ref{pipeline}.
% How to disentangle features of DR and DME is vital to improve the generalization ability \cite{chu2021learning,trauble2021disentangled,van2019disentangled,montero2020role}.
It prevents the potential decrease of generalization performance by disentangling feature representations of DR and DME grading, while exploring internal correlations between DR and DME grading.
Specifically, DETACH assigns two encoders with linear classifiers to DR and DME grading tasks.
Encoders extract latent task features $f_{DR}$ and $f_{DME}$ individually. Classifiers use the concatenation of those features, denoted as $\hat{f}_{DR}$ and $\hat{f}_{DME}$, for downstream DR and DME grading tasks, respectively.
The core of DETACH is the feature detachment, which cuts off gradient computing of features, and thus, leads to independent back-propagation streams and disentangled features of DR and DME.
For example, $\hat{f}_{DR}$, the concatenation of $f_{DR}$ and detached $f_{DME}$, will flow into the DR classifier for grading.
Thus, the DME encoder does not receive any supervision signals from the DR grading task, and vice versa. 
% Although encoders learn feature representations of DR and DME individually, however, the DETACH also explores clinical correlations between them because the classifiers utilize both those features jointly to grade DR and DME.
DETACH has the following strengths for DR and DME joint grading: 
a) encoders learn DR and DME features independently, considering different symptoms and grading criteria of DR and DME as shown in Fig.\ref{grad},
and b) classifiers explicitly explore internal correlations between DR and DME,
and c) disentangled feature representations will prevent potential bias being emphasized by entangling feature representations \cite{chu2021learning}.
% and d) physicians can reuse models for DR or DME individual grading without requiring labels of another task, which improves feasibility and flexibility.

% In summary, DETACH prevents encoders from learning entangling feature representations for DR and DME grading, whose symptoms and criteria are markedly divergent,
% meanwhile, grades them jointly, which contributes to reasonably modeling and exploring correlations between them. 
% Further, the following experiments show that our framework improves performance and generalization ability under both intra- and cross-dataset tests.

\section{Experiments}
\subsubsection{Datasets and implementation settings.}
We evaluate the proposed framework in both intra-dataset and cross-dataset tests on three fundus benchmarks: DeepDRiD\cite{DeepDRiD}, Messidor \cite{decenciere2014feedback}, and IDRiD \cite{porwal2018indian}, where the first one focuses on DR grading, and the last two have both DR and DME grading tasks. 
For the intra-dataset test, we split Messidor into five folds for cross-validation, and we use the train and test sets provided by the organizers for DeepDRiD and IDRiD. 
For the cross-dataset generalization test, we use Messidor as the train set, IDRiD as the test set, 
and group DR grade 3 and 4 in IDRiD as a new grade 3 to align DR severity levels with Messidor. 
We report the accuracy (ACC), area under the ROC curve (AUC), and macro F1-score (F1) of DR and DME grading tasks for the intra-dataset test, and additionally, report recall (REC) and precision (PRE) for the cross-dataset test. 
We adopt two ImageNet pre-trained ResNet50 as encoders, two fully connected layers with input size 4096 as linear classifiers for DR and DME, respectively, Adam as the optimizer, and set the initial learning rate as $1e^{-4}$, the batch size as 16, the dynamic difficulty-aware range as $[0.15, 1]$ with $\gamma$ decreasing in the first 400 epochs, and the training ends at the $500^{th}$ epoch.

\subsubsection{Comparisons with state-of-the-art approaches.}
% To the best of our knowledge, there is only one previous work \cite{li2019canet} explicitly exploring joint DR and DME grading, and it re-implemented two competitive multi-task learning methods, MTMR-Net \cite{liu2018mtmr} and Multi-task net \cite{chen2019multi}, suit to the joint grading task.
To the best of our knowledge, there is only one previous work \cite{li2019canet} explicitly exploring the DR and DME joint grading.
It re-modified two competitive multi-task learning methods, MTMR-Net \cite{liu2019multi} and Multi-task net \cite{chen2019multi} for the joint grading task.
We compare our proposed framework with the above approaches. 
 %following the existing implementation \cite{li2019canet}.
Note that the previous joint grading work re-modified DR grading into two categories  \cite{li2019canet}, i.e., the referable and non-referable. 
However, we consider the original fine-grained grading problem for DR and DME. 
Table \ref{iid} shows the results of the intra-dataset experiment on the Messidor and IDRiD dataset. Our method consistently outperforms other SOTA approaches under the AUC metric. 
The results indicate that our framework learns more robust feature representations.
% because we do not use extra attention mechanisms or complex modules to deal with features from encoders. 

\begin{table}[t]
\caption{Comparisons with SOTA approaches in the intra-dataset test. 
%Our method performs better against other SOTA approaches, especially in the AUC metric.
}\label{iid}
\centering
\setlength{\tabcolsep}{0.49mm}{
\begin{tabular}{@{}ccccccccccccc@{}}
\toprule
\multirow{3}{*}{Method}         & \multicolumn{6}{c}{Messidor}                                                          & \multicolumn{6}{c}{IDRID}                                                             \\ \cmidrule(l){2-13} 
                                 & \multicolumn{3}{c}{DME}                       & \multicolumn{3}{c}{DR}                        & \multicolumn{3}{c}{DME}              & \multicolumn{3}{c}{DR}               \\ \cmidrule(l){2-13} 
                                 & AUC           & F1            & ACC           & AUC           & F1            & ACC           & AUC           & F1            & ACC           & AUC           & F1            & ACC           \\ \midrule
\multicolumn{1}{c|} {CANet\cite{li2019canet}}                       & 90.5          & 66.6          & \multicolumn{1}{c|} {89.3}         & 84.8          & 61.6          &\multicolumn{1}{c|}{\textbf{71.4}} & 87.9          & 66.1          & \multicolumn{1}{c|} {78.6}          & 78.9          & 42.3          & 57.3          \\
\multicolumn{1}{c|} {Multi-task net\cite{chen2019multi}}           & 88.7          & 66.0          & \multicolumn{1}{c|} {88.8}          & 84.7          & \textbf{61.7} & \multicolumn{1}{c|} {69.4}          & 86.1          & 60.3          & \multicolumn{1}{c|} {74.8}          & 78.0          & 43.9          & 59.2          \\
\multicolumn{1}{c|} {MTMR-net\cite{liu2019multi}}                       & 89.2          & 64.1          & \multicolumn{1}{c|} {89.0}          & 84.5          & 60.5          &\multicolumn{1}{c|} {70.6}          & 84.2          & 61.1          & \multicolumn{1}{c|} {79.6}          & 79.7          & 45.3          & \textbf{60.2} \\
\multicolumn{1}{c|} {Ours} & \textbf{92.6} & \textbf{70.9} &\multicolumn{1}{c|} {\textbf{90.3}} & \textbf{86.6} & 61.6          & \multicolumn{1}{c|} {70.6}      & \textbf{89.5} & \textbf{72.3} &\multicolumn{1}{c|} {\textbf{82.5}} & \textbf{84.8} & \textbf{49.4} & 59.2          \\ \bottomrule
\end{tabular}}
\end{table}

\subsubsection{Ablation study.}
\begin{table}[t]
\caption{Ablation study in the intra-dataset test. 
%It shows both DETACH and DAW contribute to improvements in performance. 
}\label{add}
\centering
\setlength{\tabcolsep}{0.45mm}{
\begin{tabular}{@{}ccccccccccccc@{}}
\toprule
& \multicolumn{6}{c}{Messidor}   & \multicolumn{6}{c}{IDRID}    \\ \cmidrule(l){2-13} 
& \multicolumn{3}{c}{DME}        & \multicolumn{3}{c}{DR}   & \multicolumn{3}{c}{DME}   & \multicolumn{3}{c}{DR}   \\ \cmidrule(l){2-13} 
\multirow{-3}{*}{Method}         & AUC                                  & F1                                   & ACC                                  & AUC                                  & F1            & ACC           & AUC           & F1            & ACC           & AUC           & F1            & ACC           \\ \midrule
\multicolumn{1}{c|} {Joint Training}                   & 90.0                                 & 64.6                                 & \multicolumn{1}{c|} {88.4}                                 & 84.6                                 & 59.0          & \multicolumn{1}{c|} {69.7}          & 84.0          & 61.1          & \multicolumn{1}{c|} {75.7}          & 82.3          & 41.9          & 57.3          \\
\multicolumn{1}{c|} {DETACH w/ CE}                   & 91.7                                 & 69.7                                 & \multicolumn{1}{c|} {90.1}                                 & 85.5                                 & 60.2          &\multicolumn{1}{c|} {\textbf{70.7}} & 87.7          & 66.1          &\multicolumn{1}{c|} {\textbf{84.3}}         & 80.2          & 37.9          & 48.5          \\
\multicolumn{1}{c|} {Ours}  & {\textbf{92.6}} & { \textbf{70.9}} & \multicolumn{1}{c|} {\textbf{90.3}} & \textbf{86.6} & \textbf{61.6} &\multicolumn{1}{c|} {70.6}  & \textbf{89.5} & \textbf{72.3} &\multicolumn{1}{c|} {82.5} & \textbf{84.8} & \textbf{49.4} & \textbf{59.2} \\ \bottomrule
\end{tabular}}
\end{table}

To better analyze the effects of components of our proposed method, we conduct an ablation study.
Table \ref{add} shows the result of the ablation study on Messidor and IDRiD dataset in the intra-dataset test. 
The joint training is an ordinary multi-task learning model containing one ResNet50 with ImageNet pre-training and two linear classifiers for DR and DME grading. 
The performance improves with the addition of DETACH and DAW, showing the positive effect of our proposed components.
It also illustrates that the model could learn better correlations via disentangled feature representations.

\subsubsection{The cross-dataset generalization test.}
\begin{table}[t]
\caption{Comparisons and ablation study in the cross-dataset test. 
% The result shows our method is robust and generalized, compared with other SOTAs, and both DETACH and DAW contribute to a robust model with good generalization.
}\label{noniid}
\centering
\setlength{\tabcolsep}{1.27mm}{
\begin{tabular}{@{}ccccccccccc@{}}
\toprule
\multirow{2}{*}{Methods}         & \multicolumn{5}{c}{DME}                                                       & \multicolumn{5}{c}{DR}                                                        \\ \cmidrule(l){2-11} 
                                 & AUC           & F1            & ACC           & REC           & PRE           & AUC           & F1            & ACC           & REC           & PRE           \\ \midrule
\multicolumn{1}{c|} {CANet\cite{li2019canet}}                           & 83.5          & 58.2          & 78.5          & 59.8          & \multicolumn{1}{c|} {60.7}          & 78.6          & 30.2          & 45.0       & 37.5          & 36.5          \\
\multicolumn{1}{c|} {Multi-task net\cite{chen2019multi}}                & 85.5          & 56.9          & 78.5          & 59.0          & \multicolumn{1}{c|} {57.7}          & 79.2          & \textbf{32.2} & 44.6          & 38.1          & \textbf{39.0} \\
\multicolumn{1}{c|} {MTMR-net\cite{liu2019multi}}                  & 81.8          & 61.5          & 76.3          & 61.3          & \multicolumn{1}{c|} {67.3}          & 79.5          & 28.7          & 46.0          & 37.3          & 31.9          \\
\multicolumn{1}{c|} {Joint Training}               & 83.4          & 59.1          & 78.5          & 60.2          &\multicolumn{1}{c|} {61.8}          & 75.2          & 27.7          & 43.3          & 35.5          & 30.3          \\
\multicolumn{1}{c|} {DETACH w/ CE}                   & 87.3          & 64.4          & \textbf{84.3}          & 65.1          & \multicolumn{1}{c|} {68.1}          & 79.5          & 31.9          & \textbf{49.2} & \textbf{39.2} & 34.9          \\
\multicolumn{1}{c|} {Ours}  & \textbf{87.7} & \textbf{70.0} & 80.9 & \textbf{69.0} & \multicolumn{1}{c|} {\textbf{73.4}} & \textbf{79.7} & 30.8          & 42.9          & 37.0          & 36.5          \\ \bottomrule
\end{tabular}}
\end{table}

\begin{table}[t]
\caption{Comparisons on different loss functions.
%It shows DAW is suited to the grading tasks with the existence of ambiguous samples.
}\label{loss}
\centering
\renewcommand\arraystretch{1}
\setlength{\tabcolsep}{2.45mm}{
\begin{tabular}{@{}cccccccccc@{}}
\toprule
\multirow{2}{*}{Loss}         & \multicolumn{3}{c}{DME, IDRiD}                 & \multicolumn{3}{c}{DR, IDRiD}                  & \multicolumn{3}{c}{DR, DeepDRiD}                  \\ \cmidrule(l){2-10} 
                              & AUC           & F1            & ACC         & AUC           & F1           & ACC              & AUC           & F1            & ACC         \\ \midrule
\multicolumn{1}{c|} {CE}                            & 84.2          & 64.7          & \multicolumn{1}{c|} {79.6}         & 76.5          & 47.8          & \multicolumn{1}{c|} {57.3}          & 83.1          & 51.9          & 61.0          \\
\multicolumn{1}{c|} {FL \cite{lin2017focal}}         & 86.2          & 65.4          & \multicolumn{1}{c|} {80.6}          & 75.3          & 40.5          &\multicolumn{1}{c|} {\textbf{58.3}} & 83.2          & \textbf{53.0} & \textbf{63.0} \\
\multicolumn{1}{c|} {GCE \cite{zhang2018generalized}}               & 88.6          & 71.3          &\multicolumn{1}{c|}  {80.6}          & 77.4          & 40.3          & \multicolumn{1}{c|} {52.4}          & 82.9          & 52.2          & 60.0          \\
\multicolumn{1}{c|} {$\mathcal{L}_{DAW}$} & \textbf{90.2} & \textbf{74.2} &\multicolumn{1}{c|} {\textbf{83.5}} & \textbf{82.9} & \textbf{50.2} & \multicolumn{1}{c|} {55.3}          & \textbf{84.6} & 52.0          & 60.1          \\ \bottomrule
\end{tabular}}
\end{table}

A recent study suggests evaluating models in the cross-dataset test to measure generalization capacity \cite{geirhos2020shortcut}. 
To further investigate the generalization and robustness of our method, we conduct experiments with the above approaches in the cross-dataset test. 
Table \ref{noniid} shows the results of those approaches and ablation study under the cross-dataset test setting. 
The results show that our method is more robust and generalized than existing approaches.
Besides, the considerable gap between the joint training and DETACH implies our disentangled learning strategy has better generalization ability than entangling feature representations. 
% It further verifies the assumption in our paper, i.e., entangling representations may emphasize bias in the joint grading task and thus harm generalization performance.

\subsubsection{Loss study.}

Finally, to verify the effectiveness of DAW,  we conduct experiments comparing it with three popular loss functions, including Cross-Entropy Loss (CE),  Focal Loss (FL) \cite{lin2017focal}, and Generalized Cross-Entropy Loss (GCE) \cite{zhang2018generalized}.
To avoid the influence of entangling feature representation in joint grading, we evaluate the above loss functions by individual DR or DME grading tasks on the IDRiD and DeepDRiD.
We adopt a ResNet50 as the backbone and a fully connected layer with input size 2048 as the classifier, set the training epochs as 1000, and the batch size as 8.
We adopt default parameter settings for the above loss functions for fairness, i.e., $\gamma = 2$ for FL, $q = 0.7$ for GCE, and dynamic difficulty-aware range $[0,1]$ for DAW.
The result in Table \ref{loss} illustrates the superiority of DAW on the grading task, where difficult samples require a reasonable learning strategy.
This result also implies that DAW helps the model learn robust feature representations for grading tasks.
% The uniform dynamic difficulty range $[0,1]$ across different grading tasks shows the effectiveness and robustness of our method.
% and empirically, researchers could tune this range to get better results in public and private datasets.

\section{Conclusion}
In this paper, we focus on the joint grading task of DR and DME, which suffers from difficult samples and potential generalization issues.
To address them, we propose the dynamic difficulty-aware weighted loss (DAW) and dual-stream disentangled learning architecture (DETACH).
DAW measures the difficulty of samples adaptively and learns from simple to difficult samples dynamically.
DETACH builds disentangled feature representations to explicitly learn internal correlations between tasks, which avoids emphasizing potential bias and mitigates the generalization ability degeneration.
We validate our methods on three benchmarks under both intra-dataset and cross-dataset tests. 
% to verify the effectiveness of our methods.
Potential future works include exploring the adaptation of the proposed framework in other medical applications for joint diagnosis and grading.
% Our framework achieves new SOTA results on DR and DME joint grading.

\subsubsection{Acknowledgments.} 
This work was supported by funding from Center for Aging Science, Hong Kong University of Science and Technology, and Shenzhen Science and Technology Innovation Committee (Project No. SGDX20210823103201011).

\bibliographystyle{splncs04}
\bibliography{paper492}
\end{document}